\documentclass[letterpaper, 10 pt, conference]{ieeeconf}  

\IEEEoverridecommandlockouts                              

\usepackage{graphics} 
\usepackage{epsfig} 
\usepackage{balance}
\usepackage{cite}
\usepackage{stmaryrd}
\usepackage{hyperref}

\bstctlcite{IEEEexample:BSTcontrol}

\begin{document}

\title{\LARGE \bf
Generating Natural and Expressive Robot Gestures through Iterative Reinforcement Learning with Human Feedback using LLMs
}

\author{Chris Lee$^{1}$, Flora Salim$^{1}$, Benjamin Tag$^{1}$, Francisco Cruz$^{1,2}$
\thanks{$^{1}$Chris Lee, Flora Salim, Benjamin Tag and Francisco Cruz are with the School of Computer Science and Engineering, University of New South Wales, Sydney, Australia. Corresponding Author:
        {\tt\small christopher.lee1956@student.unsw.edu.au}}%
\thanks{$^{2}$Francisco Cruz is with Universidad Central de Chile, Santiago, Chile}%
}

\maketitle
\thispagestyle{empty}
\pagestyle{empty}

\begin{abstract}

Expressive gestures are essential for natural and effective communication, complementing speech when verbal cues alone are insufficient (e.g., pointing). For social robots such as the humanoid Pepper, producing natural and expressive movements is critical for improving human-robot interaction (HRI) and long-term acceptance. However, generating gestures remains challenging due to reliance on expert-authored animations, resulting in rigid behaviors that are impractical for dynamic and diverse environments. Alternatively, machine learning approaches often struggle to capture perceived naturalness, becoming increasingly challenging with more degrees of freedom. Consequently, producing expressive robot gestures requires a system that can adapt to the environment while adhering to social norms and physical constraints. Recent advances in large language models (LLMs) enable dynamic code generation, offering new opportunities for runtime gesture synthesis from natural language. In this paper, we integrate ChatGPT into the humanoid robot Pepper to generate co-speech gestures aligned with conversational output. While this baseline enables flexible gesture generation, the resulting motions are often perceived as stiff and unnatural. To address this limitation, we introduce an iterative reinforcement learning with human feedback (RLHF) system that finetunes gesture generation based on user evaluations, leveraging an iterative user study to compare Pepper’s generated gestures. Our results show that RLHF improved the LLM’s co-speech generative capabilities, producing more expressive, relevant and fluid movements. 

\end{abstract}

\section{INTRODUCTION}
\begin{figure*}[hbt!]
  \centering
  \includegraphics[width=0.9\linewidth,keepaspectratio, trim={0cm 2cm 0cm 2.4cm}, clip]{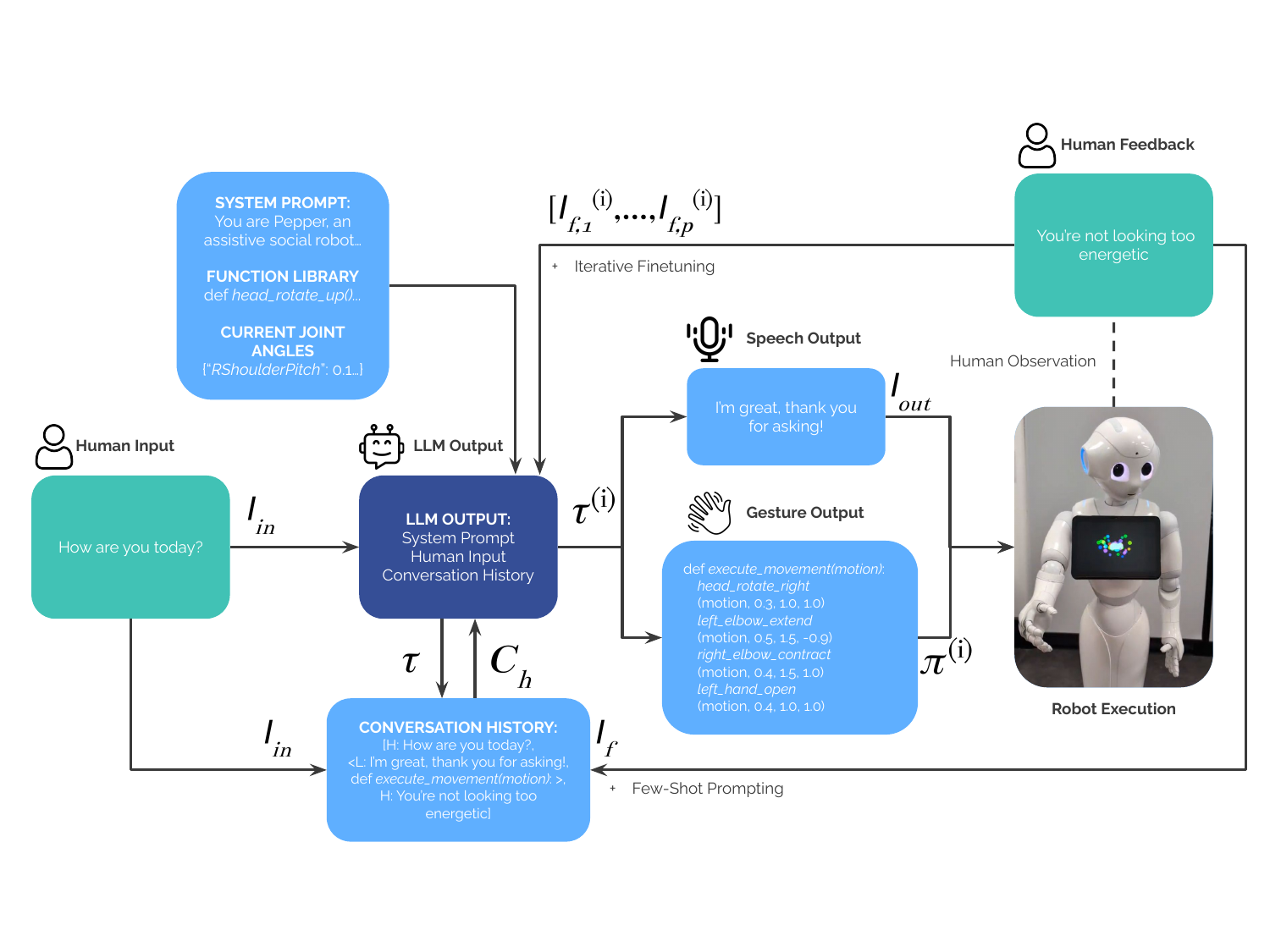}
  \caption{Overview of expressive co-speech gesture generation system with iterative reinforcement learning with human feedback.}
  \label{fig:figure1}
\end{figure*}

The rise of generative artificial intelligence (AI) has opened new opportunities for robotic systems to adapt and operate in dynamic, real-world environments. As robots are increasingly deployed in socially interactive settings, they require learning mechanisms that go beyond static pretraining and hand-crafted design. While current robotics research primarily targets functional tasks such as manipulation, navigation, and speech-based communication~\cite{zhang_large_2023,zhang_generative_2025,zeng_large_2023,wang_large_2024},
domains that are well-defined and objectively measurable, socially expressive behaviors like gesture generation remain challenging due to their subjective and perception-driven nature~\cite
{yoon_speech_2020,zabala_modeling_2022}. Ultimately, to enable social robots that are not only functional but also preferred by users, robotic systems must therefore be capable of continuously learning from human feedback.

The performance of social robots is often evaluated using subjective human ratings, such as the GODSPEED questionnaires~\cite{bartneck_measurement_2009}, which introduce personal bias and abstraction. However, this subjectivity underscores the necessity of human feedback when optimizing social behaviors, as improvements in functional metrics do not necessarily correspond to users’ perceptions of intelligence or interaction quality~\cite{hoffman_primer_2021}. Perceived qualities such as naturalness or smoothness cannot be fully captured by predefined or task-driven metrics. Therefore, developing anthropomorphic and likable social robots requires integrating learning mechanisms that adapt directly to communicated human preferences, such as interactive reinforcement learning.

Supporting users with varying robotics expertise is also crucial, as they interpret robot behavior differently~\cite{hoffman_primer_2021}. Intuitive human-robot interaction (HRI) often relies on natural language~\cite{barmann_incremental_2024}, however, non-verbal cues such as gestures, perceived emotion or proxemics can serve as implicit reward signals~\cite{akalin_reinforcement_2021,kaufmann_survey_2025}
. This motivates the use of large language models (LLMs), whose ability to incorporate human feedback or examples (e.g., few-shot learning)~\cite{barmann_incremental_2024,ouyang_training_2022},
can help align robot behavior with human preferences.

The rapid growth of LLM research has enabled robotics to integrate multiple modalities~\cite{zhang_large_2023,zeng_large_2023} and generate motion code in real time~\cite{zhang_generative_2025,mahadevan_generative_2024,vemprala_chatgpt_2023,sobo_evaluating_2025}. 
In embodied agents, their modular structure allows coordinated processing of multimodal data and environmentally grounded actions~\cite{wang_large_2024,zeng_large_2023}, while richer feedback modalities can capture more nuanced user preferences~\cite{de_heuvel_impact_2025}. Despite these advantages, challenges remain in contextual grounding, embodiment, and real-time adaptation, and reliance on offline pretraining can introduce cost and dataset rigidity bottlenecks~\cite{zhang_large_2023,zeng_large_2023,zhang_generative_2025}. 
Consequently, robots operating in real-world environments require the ability to learn and adapt dynamically, rather than depending solely on predefined or curated knowledge.

In this paper, we investigate GPT-4 for generating expressive co-speech gestures on the Pepper robot~\cite{pandey_mass-produced_2018}. Initial LLM-generated motions were often stiff and unnatural, potentially limiting user acceptance if used in an HRI experiment. To address this, we conducted an iterative online user study in which participants rated videos of Pepper performing expressive co-speech gestures aligned with spoken phrases. Participants’ ratings were then used as feedback to finetune the LLM’s gesture code generation, allowing us to evaluate whether this iterative process improves perceived expressiveness. Hence, we introduce an iterative reinforcement learning with human feedback (RLHF) system (Fig.~\ref{fig:figure1} that finetunes gesture generation based on user evaluations. This leads to our research questions; \textbf{RQ1:} Can human feedback be used to finetune an LLM’s code generation to improve the perceived expressiveness of generated robot gestures? \textbf{RQ2:} To what extent can LLMs generate code for expressive co-speech gestures on a humanoid robot?

The online user study had five iterations, finetuning GPT-4 a total of four times, each with new human feedback. Our work demonstrates that LLMs can be finetuned in both offline and online settings to learn human preferences from subjective feedback. By coupling the model with a library of low‑level motion primitives, we enable it to generate gesture code that can adapt flexibly to new contexts and embodiment constraints. Integrating human feedback directly into the reinforcement learning loop further allows the model to be steered toward increasingly expressive and human‑aligned robot gesture code over time.

\section{RELATED WORKS}
\subsection{Expressive Motion and Social Behavior in Humanoid Robots}
In real-world settings, verbal communication is often ambiguous, noisy, or incomplete, making non-verbal behaviors such as body language and gestures critical for conveying intent and meaning~\cite{mehrabian_nonverbal_2017}. 
In HRI, gestures like pointing, can improve robot legibility and user understanding, leading to shorter interactions, faster error detection, and fewer overall errors~\cite{breazeal_effects_2005}.  
However, gestures must be used carefully, as they can become confusing without sufficient contextual grounding~\cite{ende_human-centered_2011}, especially when they are symbolic or metaphorical. Overall, gestures are essential for HRI performance, as they shape task efficiency, error rates, and users’ interpretation of robot intent. 

Beyond functional roles, expressive movements convey affective and social information, shaping perception, engagement, and trust during interaction. Such movements can signal responsiveness to nearby people~\cite{lombardi_would_2025} and communicate emotional cues that language alone cannot express~\cite{ng-thow-hing_synchronized_2010}. 
Because these cues directly influence how humans interpret and relate to robots, any system that autonomously learns and generates expressive motions must ultimately be aligned with human perception and preferences.

\subsection{Reinforcement Learning with Human Feedback}
As social robots take on increasingly complex and long-term interactions, they must be able to learn incrementally from human feedback, often provided by non-expert users~\cite{akalin_reinforcement_2021}. However, gathering feedback can be costly at large scales and when factoring fatigue. Designing RLHF remains challenging, particularly in shaping reward signals that reflect human intent while remaining practical for learning~\cite{moreira_deep_2020, balsells_autonomous_2023, kaufmann_survey_2025}. Early approaches like TAMER relied on scalar feedback or structured mappings from verbal advice to reward variables~\cite{knox_interactively_2009, kuhlmann_guiding_2004}, while pairwise preference comparisons later enabled RLHF to scale to deep reinforcement learning~\cite{christiano_deep_2017}. 
More recent work has explored implicit feedback signals~\cite{wang_affective_2022}
and learned reward models~\cite{juan_shaping_2021}, as well as leveraging LLMs to support richer and more scalable human feedback~\cite{ouyang_training_2022, jin_data-efficient_2023}. 

Integrating LLMs with RL combines their strengths as LLMs provide broad pretrained knowledge, while RL enables contextual grounding through environmental interaction, with popular models already applying RLHF~\cite{cao_survey_2025}. This synergy benefits robotics by supporting multimodal understanding, multitask generalization, efficient learning, and flexible reward shaping~\cite{cao_survey_2025,kaufmann_survey_2025,wang_reinforcement_2024}. LLMs can also substitute human feedback with AI-generated feedback in self-rewarding systems~\cite{kaufmann_survey_2025,wang_reinforcement_2024,yuan_self-rewarding_2024},
that ultimately improves scalability but weakens human-alignment.

Human interpretability and alignment are crucial in critical settings, such as robots deployed in  medical institutions, making traditional RL potentially unsafe or inefficient and the environment difficult to reproduce~\cite{balsells_autonomous_2023}. Interactive RL is particularly effective in early learning stages, where corrective feedback often provides more guidance than evaluative rewards~\cite{moreira_deep_2020}. Human feedback can be incorporated via offline and online learning (e.g., i-sim2real and SKiP), the offline phase stabilizes behavior while improving long-horizon performance~\cite{abeyruwan_i-sim2real_2023,wang_skill_2022}. While offline finetuning maximizes performance on fixed data, online learning is essential to maintain alignment with evolving human preferences~\cite{stiennon_learning_2020}.
Iterating between these phases enables more efficient learning by mitigating early-stage exploration failures and adapting to dynamic user expectations. 

In robot motion generation, additional constraints further limit safe exploration. For instance, SEED~\cite{hiranaka_primitive_2023} learns high-level primitives like reaching and grasping rather than low-level joint commands, highlighting the need to align human feedback with the action space. Combining pretrained LLM knowledge with RLHF’s sample efficiency offers a promising approach for learning expressive, human-aligned robot low-level motions iteratively.

\subsection{LLM based code generation and robotic code}
GenAI for robot code spans outputs from low-level movement functions to high-level policies, with higher-level abstractions generally being more human-aligned due to their explicit intent. A key challenge is enabling GenAI to compose low-level functions that reflect high-level intent.
While larger datasets can improve generation quality, collecting human-aligned data at scale is costly, motivating approaches that learn preferences efficiently. Prompt-based methods are commonly used to inject human feedback into LLM-driven robot code generation~\cite{tarakli_interactive_2024,huang_emotion_2025,barmann_incremental_2024,mahadevan_generative_2024}, with systems such as ECLAIR, GenEM++, and B\"armann et al.'s framework focusing on high-level actions (e.g., pick-and-place) rather than low-level motion. 
This exposes a gap in using GenAI and human feedback to generate expressive, unconstrained low-level robot motions for co-speech gesture generation.

Creating robot gestures is inherently non-deterministic, as multiple gestures may be appropriate for the same input, particularly in co-speech contexts, making objective evaluation challenging~\cite{lu_co-speech_2023,yazdian_gesture2vec_2022}.
In prolonged interactions, it is natural and desirable for robots to avoid repetitive behaviors~\cite{zabala_modeling_2022}, and LLMs should also be capable of generating diverse co-speech gesture code~\cite{sobo_evaluating_2025}. 
Quantitative and objective evaluations struggle to capture human preference across non‑deterministic generative outputs. Likewise, although low‑level primitive movement functions offer adaptability and generalization, they do not inherently encode the expressive, human‑aligned patterns seen in higher‑level movement policies. These limitations make iterative human feedback especially important for steering the model toward producing appropriate and expressive co‑speech gestures.

\begin{figure*}[hbt!]
  \centering
  \includegraphics[width=\linewidth,height=0.45\textheight,keepaspectratio, trim={0.5cm 0cm 2.9cm 0.2cm}, clip]{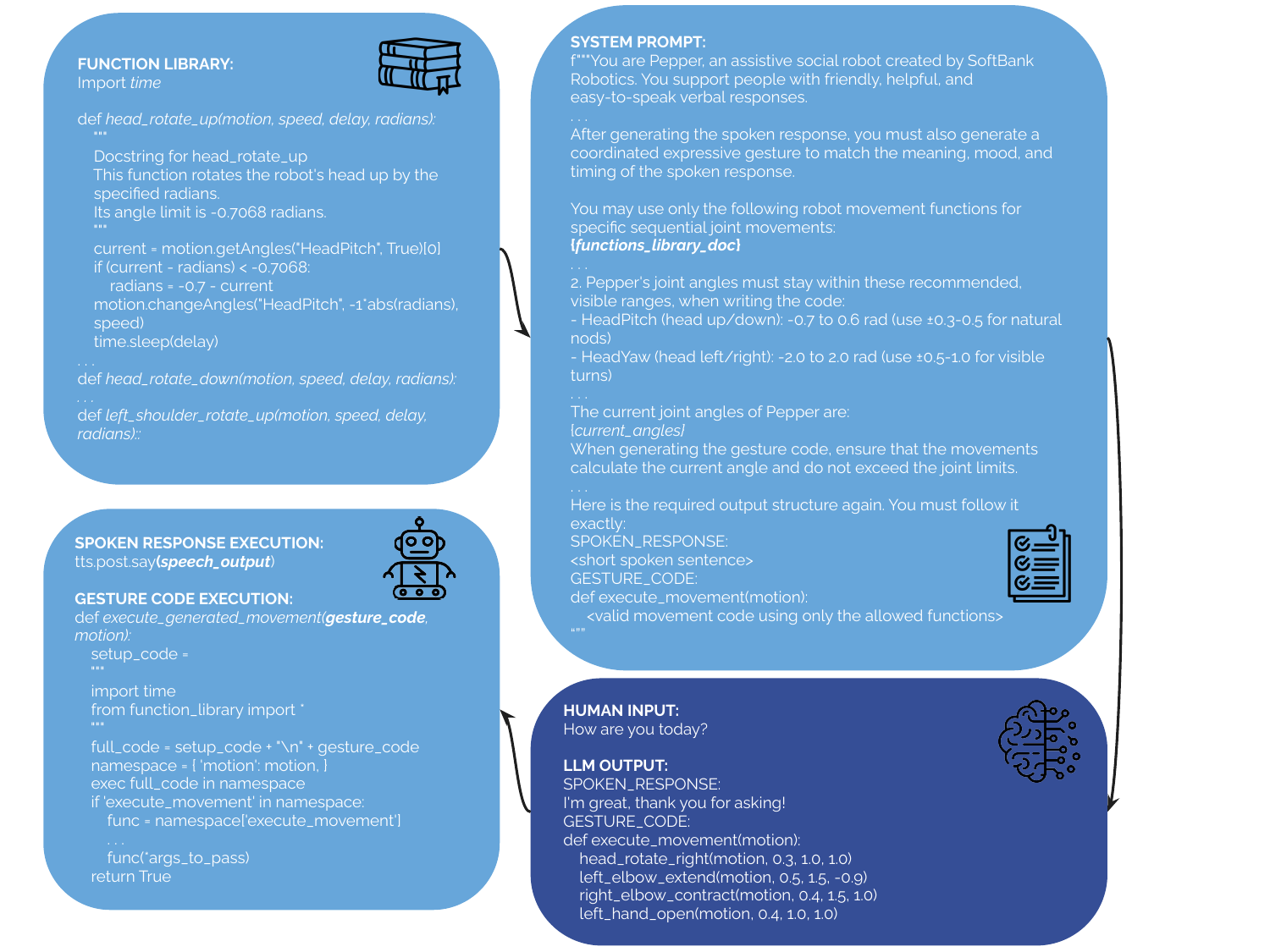}
  \caption{Pipeline from function library of primitive robotic motions to final execution.}
  \label{fig:figure2}
\end{figure*}

\section{METHODOLOGY}

\textbf{Overview:} Our RLHF system inherits the procedural structure of recent gesture generation processes~\cite{mahadevan_generative_2024,roy_gpt-driven_2025,huang_emotion_2025}, using them to generate co-speech gestures that cannot be evaluated functionally, leveraging human evaluation to optimize their expressiveness. As seen in Fig.~\ref{fig:figure1}, it takes natural language input $l_{in} \in \mathcal{L}$ and generates a co-speech tuple $\tau = \langle l_{out},\pi\rangle$ of conversational output $l_{out} \in \mathcal{L}$ and a continuous sequence of parameterized primitive robotic functions, with length $s$ to compose the expressive gesture $\pi = [a_{1},\ldots,a_{s}]$. We denote the sequence of parameterized functions as a temporally extended policy $\pi$, with each function as an action $a$.

Each policy will then be subject to human evaluation, elaborated in the user study section, to act as the reward for finetuning. Only the motion sequence will be finetuned and not the conversational output, while the co-speech tuple is generated by OpenAI's GPT-4.1 (gpt-4.1-2025-04-14). GPT has strengths in general purpose and linguistic understanding~\cite{zeng_large_2023}, suited for co-speech and our prompt structure enforces safe code generation. The system will also keep a \textit{conversation history} $C_{h} = [l_{in,1},\tau_{1},\ldots,l_{in,c},\tau_{c}]$ where $c$ is the number of interaction cycles between person and robot. $C_{h}$ is injected into the prompt to give the LLM contextual understanding of previous interactions with the person. 

\textbf{Motion Sequence Generation:} To ensure that the LLM-generated motion code is executable on the Pepper robot under the NAOqi v2.5 framework, we adhere to the official Aldebaran documentation\footnote{Pepper Aldebaran Documentation: \url{http://doc.aldebaran.com/2-4/home_pepper.html}} specifying joint names and angle limits. We implement a library of parameterized low-level joint motion functions with docstrings, which is injected into the LLM prompt as a structured knowledge base, enabling the model to generate properly embodied motions in real time from natural language input (Fig.~\ref{fig:figure2}). Each primitive function corresponds to a single joint (e.g., \textit{RShoulderPitch}) and specifies the target angle in radians, execution speed, and a post-motion delay. Temporal coordination across the motion sequence is achieved by encouraging the LLM to insert explicit timing intervals between primitives, enabling synchronized co-speech gesture generation. Altogether, the sequence of reusable primitive functions and timing intervals will compose $\pi$.

\textbf{LLM Prompting:} 
By default, when prompting GPT to generate expressive co-speech gestures alongside dialogue, the base model will lack embodiment and environmental awareness. We address this through a system prompt (Fig.~\ref{fig:figure2}) that constrains the LLM to produce safe, executable robot code while adopting a personality aligned with Pepper’s role as a social robot. As co-speech gestures are non-functional, the injected state information is limited to Pepper’s joint limits, represented via a library of primitive motions, and its current joint angles. We used \textit{role-playing} prompts to define Pepper’s conversational persona (e.g., \textit{“You are Pepper, an assistive social robot”}) and \textit{directional} prompting to guide natural, polite, and concise speech (e.g., \textit{"Use simple words when possible, stay positive and patient"}), enabling consistent and human-appropriate conversational output.

As co-speech gestures can have multiple motions relevant for the same utterance, we set the temperature at 1.5 despite other implementations being close to 0~\cite{roy_gpt-driven_2025,huang_emotion_2025}, to encourage expressive variability. Despite this increased stochasticity, strict output schemas were maintained through \textit{constraint-based code generation} (e.g., \textit{"You must NOT include imports, comments...", "All motion function calls must follow this exact parameter order"}), \textit{instruction} prompting (e.g., \textit{"STRICT OUTPUT RULES - YOU MUST FOLLOW ALL OF THEM"}), and \textit{program synthesis} that ensures coherent primitive motion sequences. Prompt robustness was further improved using \textit{meta-prompting}, in which the system prompt was iteratively refined using LLM-assisted analysis. Additionally, for contextual grounding and real-time adaptation, we applied \textit{few-shot prompting} by maintaining a \textit{conversation history} of previous user interactions. 

\textbf{Iterative Human Feedback:}  Similar to EMOTION++~\cite{huang_emotion_2025}, we incorporate human feedback through an iterative combination of offline and online learning, where each iteration $i \in [1,2,\ldots,i_{max}]$ will receive human feedback $l_{f}$ to reproduce reinforcement learning (Fig.~\ref{fig:figure1}). Each $i$ will use the same set of $l_{in}$ with length $n$ to produce corresponding co-speech tuples $[l_{in,1}^{(i)},\ldots,l_{in,n}^{(i)}]\mapsto[\tau_{1}^{i},\ldots,\tau_{n}^{i}]$. However, while EMOTION++ uses their research team for feedback, we crowdsource our human evaluations from an online user study to have a diversified understanding of  human preferences and also scalability~\cite{balsells_autonomous_2023,roy_gpt-driven_2025}. 

We collect several evaluations on the same motion sequence generated by the LLM to allow us to mimic episodic learning in RL, which is difficult to reproduce in HRI~\cite{balsells_autonomous_2023}. Every gesture $\pi^{(i)} \in [\pi_{1}^{(i)},\ldots,\pi_{n}^{(i)}]$ will receive $l_{f} \in [l_{f,1},\ldots,l_{f,p}]$, where $p$ is the number of participants, creating a total of $n*p$ samples of $l_{f}$. For offline finetuning, the LLM learns to predict $l_{f} \mapsfrom \langle \pi^{(i)}_1, \pi^{(i)}_2, l_{out}\rangle$ by using \textit{direct preference optimisation (DPO)} and pairwise comparison, identifying which of the two candidate gestures $\pi^{(i)}_1$ or $\pi^{(i)}_2$ produces higher $l_{f}$ and is preferred by participants. $l_{out}$ remains the same for each iteration of $\tau^{i}$. Meanwhile, online finetuning will have the LLM summarize $l_{f,sum} \mapsfrom [l_{f,1},\ldots,l_{f,p}]$, to append to the \textit{conversation history} $C_{h}$ injected into the prompt as \textit{few-shot} when given $l_{in}$. Each subsequent $i$ will retain $C_{h}$ from previous iterations, where each $i$ will have $c = i-1$. To analyze the effectiveness of our system, the human evaluations of every $i \in [1,2,\ldots,i_{max}]$ will be compared.

\section{USER STUDY}
\subsection{Overview}
To investigate how human feedback enhances LLM-based motion sequence generation for expressive co-speech gestures, we conducted an iterative online user study\footnote{The full survey instrument preview can be viewed online at:\\
\url{https://unsw.au1.qualtrics.com/jfe4/preview/previewId/df7d60a3-d99b-4049-bee7-6d6237f599d5/SV_4Zrg5aTNYYUI2qy?Q_CHL=preview&Q_SurveyVersionID=current}} organized into five evaluation rounds, where $i_{max} = 5$. In each iteration, participants were shown video recordings of the Pepper robot speaking and gesturing, and asked to evaluate the generated co-speech gestures. Across all iterations, the verbal output remained fixed, while feedback collected in each round was used to fine-tune the LLM and refine gesture generation for subsequent iterations. This iterative design isolates the effect of human feedback on low-level gesture generation and enables a systematic analysis of how expressive co-speech behaviors evolve over time, despite the inherent non-determinism and subjectivity of gesture interpretation. This user study addresses the research questions introduced earlier by comparing human evaluations across iterations, and we hypothesize for \textbf{RQ1} that: \textbf{H1:} Co-speech gestures produced in the final iteration will be perceived as more satisfactory than those generated in the first iteration.

\subsection{Participants}
Participants were recruited via the online Prolific\footnote{\href{https://www.prolific.com/}{Prolific Website Link: https://www.prolific.com/}} platform, widely used for academic and industry research studies. We prioritized native English speakers as inclusion criterion and recruited participants from AU, the UK, and the US, as the LLM verbal output was in English in the recorded videos. To mitigate language barriers when evaluating co-speech gestures, non-native speakers were also allowed, with English proficiency self-determined. A total of 250 adults (18+) participated across five study iterations of 50 participants each, targeting the general public to capture diverse perspectives on perceived expressiveness. Nonetheless, cultural and language biases remain due to the use of English stimuli. Ethics approval\footnote{UNSW HRECS Office - Reference No. iRECS8045} was obtained from the authors’ institution prior to data collection.

\subsection{Survey Procedure}
The online survey was hosted on the Qualtrics platform and deployed five times, each with a new cohort of participants. The participants are linked the documentation of the Pepper robot's joints, to familiarize themselves with its movement capabilities and limits. This ensures that no feedback is given where the movements are physically unachievable. In each deployment, participants viewed a randomized subset of 20 short videos showing the Pepper robot performing generated co-speech gestures. After each video, participants rated the perceived \textit{expressiveness}, \textit{relevance}, and \textit{fluidity} of the gesture using a five-point Likert scale (1 = “Not at all”, 5 = “Very much”). Participants could also provide optional free-text comments (e.g., \textit{“arms barely moved”}) to capture more natural feedback. Prior to each study iteration, co-speech gestures were generated using the LLM from a fixed set of phrases and recorded through Pepper. The first iteration used the base GPT-4.1 model, while each subsequent iteration fine-tuned the model from the previous round using the collected human feedback.

A total of 50 distinct phrases were used in the study, grouped into five categories with 10 phrases each: \textit{greeting, encouragement, apology, explanation,} and \textit{collaboration}. Phrases were generated using GPT-4.1 with the same social embodiment prompting guidelines described in the methodology, ensuring reproducibility in live deployment settings. The phrases were designed to reflect casual, everyday speech (e.g., \textit{“Good job, keep it up!”, “Hi, how are you doing today?”,} and \textit{“Sorry if that caused any confusion”}). Semantic overlap was intentionally introduced across phrases to encourage non-deterministic co-speech gesture generation for similar meanings. In each iteration, 1,000 feedback samples were collected, with each phrase-gesture pair receiving 20 evaluations. To ensure balanced exposure across phrase categories, the full set was divided into five subsets, each containing two phrases from every category. Participants evaluated two randomly selected subsets, resulting in an equal distribution of phrase types across the study.

\begin{figure*}[hbt!]
  \centering
  \includegraphics[width=0.95\linewidth,keepaspectratio, clip]{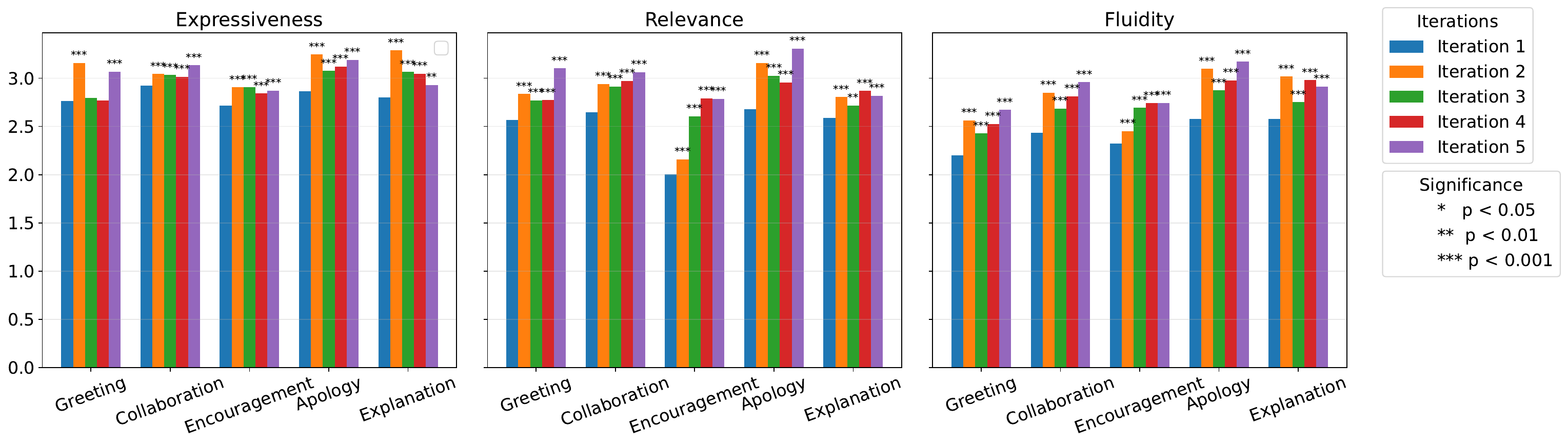}
  \caption{Perceived average expressiveness, relevance, and fluidity user ratings across categories per iteration. Although Iteration 1 shows a significant improvement, subsequent iterations fluctuate downward before stabilising and then improving consistently through Iteration 5.}
  \label{fig:figure3}
\end{figure*}

After each survey deployment, except the final iteration, collected human feedback was used to finetune the LLM. Offline finetuning via the OpenAI API trained the model to learn preferred gesture code sequences for semantically similar phrases, reinforcing patterns linked to greater perceived expressiveness, relevance, and fluidity For pairwise comparison we permutated within the five categories, then had a separate comparison for each combined participant average rating leading to a total of 1350 gesture code comparisons. Another 150 comparisons between iterations were included. In parallel, online finetuning emulated conversational feedback, again using combined participant averages, and free-text comments were summarized into a single feedback description using the LLM. This consolidated feedback, together with the phrase and generated code, was appended to the conversation history $C_{h}$, as \textit{few-shot} prompting examples. 
Both offline and online finetuning accumulated feedback across iterations and by the final iteration, the model had been trained on 4000 feedback samples while the conversation history contained $c = 4$ prior feedback cycles.

\section{RESULTS \& DISCUSSION}
\subsubsection{Quantitative Results}
Using Welch’s unpaired t-test, we evaluate \textbf{H1}, testing whether perceived expressiveness, relevance, and fluidity improve across iterations of our RLHF system. For each iteration, user ratings were first averaged per phrase and then aggregated across phrase categories. As shown in Fig.~\ref{fig:figure3}, Iteration 5 exhibits a statistically significant improvement over Iteration 1 across all metrics and categories (p\textless0.001), ultimately supporting \textbf{H1}. Most intermediate iterations also show statistically significant gains relative to the baseline. Notably, Iteration 2 demonstrates the largest immediate improvement, highlighting the strong effect of incorporating human feedback through few-shot prompting. Subsequent iterations exhibit partial regression before stabilizing, suggesting a shift from early rapid adaptation toward more constrained and incremental refinement.


\begin{figure}[hbt!]
  \centering
  \includegraphics[width=0.9\linewidth,keepaspectratio, clip]{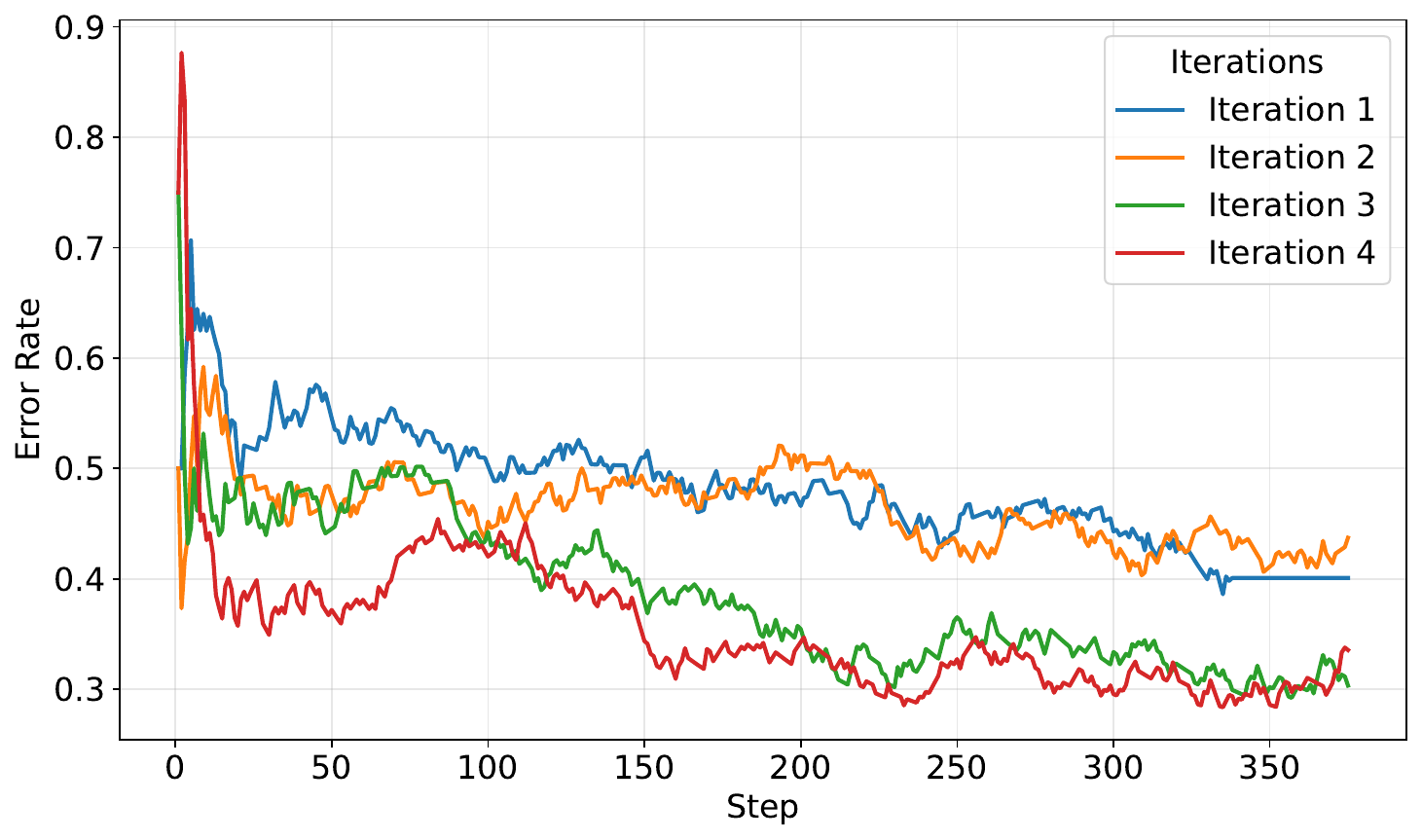}
  \caption{Smoothed Error Rate per iteration during DPO finetuning.}
  \label{fig:figure6}
\end{figure}

Relevance and fluidity improved consistently across iterations, exhibiting larger gains than expressiveness (e.g., Encouragement–Relevance increased from 2.00 to 2.78). In contrast, the largest overall improvement in expressiveness was more modest ($\Delta$ = 0.32 for Apology–Expressiveness), likely reflecting the more abstract and subjective nature of expressiveness compared to the more functional dimensions of relevance and fluidity. Apology showed the strongest overall improvement across all metrics, with Greeting and Encouragement also demonstrating substantial gains. This suggests that gestures associated with clear semantic intent and salient expressive cues were more readily reinforced through human feedback, whereas more ambiguous categories such as Collaboration and Explanation required additional learning to achieve comparable improvements. Overall, our results address \textbf{RQ2} by showing positive attitudes towards gesture quality with roughly half of the fifth iteration evaluations exceeding 3 out of 5.

\begin{figure*}[hbt!]
  \centering
  \includegraphics[width=0.95\linewidth,keepaspectratio, clip]{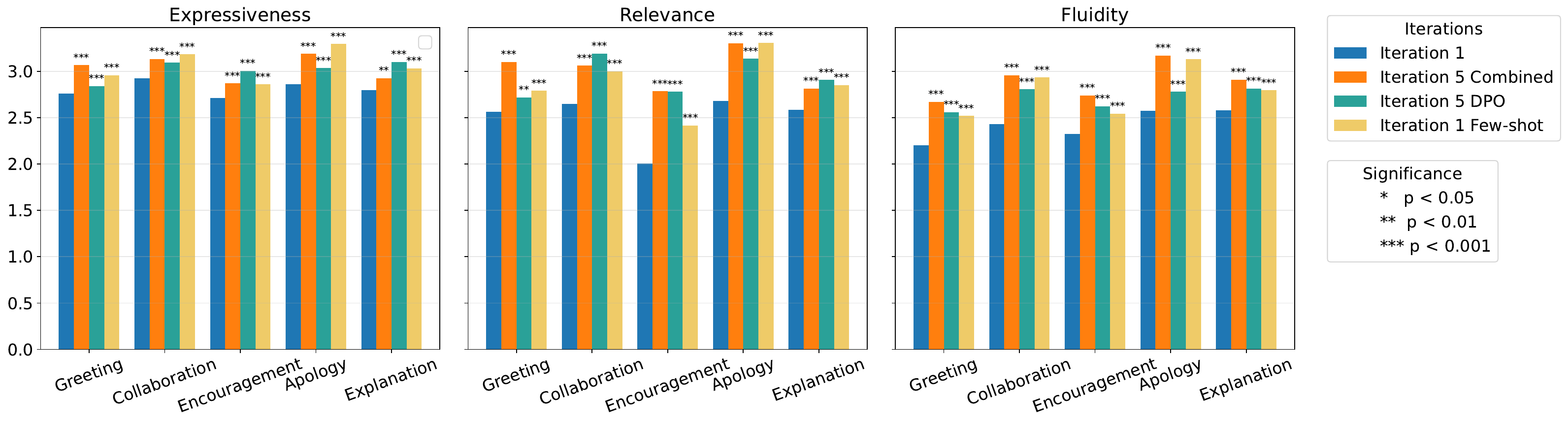}
  \caption{Perceived average expressiveness, relevance, and fluidity user ratings across categories under ablation testing.}
  \label{fig:figure4}
\end{figure*}

\begin{figure}[hbt!]
  \centering
  \includegraphics[width=0.95\linewidth,keepaspectratio, clip]{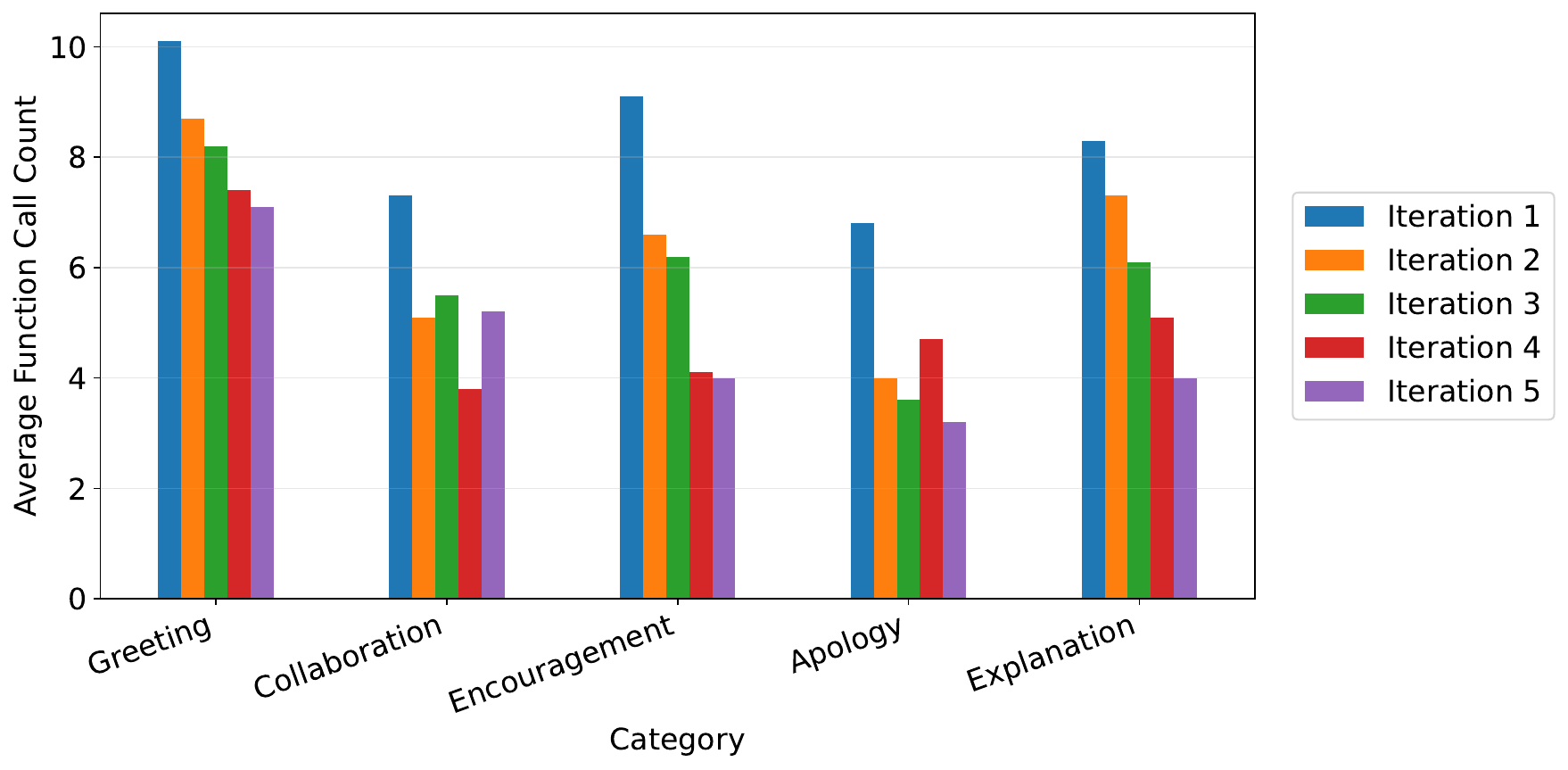}
  \caption{Average primitive movement functions called per co-speech gesture sequence across categories per iteration.}
  \label{fig:figure7}
\end{figure}

We conducted additional ablation experiments (Fig.~\ref{fig:figure4}) to assess the relative contributions of few-shot prompting and DPO. We compared the base model from Iteration 1 with the Iteration 5 combined system, alongside two ablations, removing few-shot prompting from the Iteration 5 model and injecting the few-shot conversation history into the base model. Error rates decreased across iterations (see Fig.~\ref{fig:figure6}), indicating that the model learned user preferences, however, as observed in the qualitative analysis, this learning alone often produced robust yet unnatural gesture patterns. Few-shot prompting, both in the base and combined models, yielded the largest overall gains, alternately improving expressiveness. In contrast, the contribution of DPO was most evident in relevance and fluidity, where the combined model consistently outperformed the base model, suggesting that iterative offline fine-tuning enabled the LLM to internalize and prioritize relevant primitive motion-level traits associated with smoother and more coherent gesture execution.


\subsubsection{Qualitative Gesture Generation Changes}
Across iterations, we observed systematic changes in co-speech gesture generation in response to human feedback. In the \textit{first iteration}, gestures showed limited diversity, dominated by repetitive arm-raising and open-hand movements, reflecting early exploration for reward signals. The \textit{second iteration} demonstrated the strong influence of few-shot feedback in constraining behavior, where comments discouraging arm use led the model to largely suppress arm motion. Similarly, the \textit{third iteration} further reduced gesture complexity, with the model favoring head nods over full-body gestures for high-context utterances, suggesting a conservative strategy under semantic ambiguity. In the \textit{fourth iteration}, gesture categories began to converge, with pruning of redundant functions (see Fig.~\ref{fig:figure7}) and improved fluidity, alongside the emergence of novel behaviors (e.g., forward torso bends during apologies). Arm movements became slower and more deliberate, though head motion remained abrupt.

In the \textit{fifth iteration}, gesture diversity declined in the \textit{combined model}, where few-shot prompting encouraged reuse of motion sequences from earlier iterations. This reuse of prior gestures suggests that the LLM captures which gestures are preferred, but struggles to reliably embody these preferences in executable robot motion. When \textit{conversational history was removed}, the model exhibited severe mode collapse, reusing the same four to five gestures across all categories. This suggests that DPO alone was insufficient to translate semantic preference into diverse motion code. In contrast, using the full \textit{conversation history with the base GPT model} produced similar gestures to the combined model and occasionally novel motions, albeit with reduced fluidity. Overall, these observations indicate that human feedback effectively prunes inappropriate gestures and enables limited novelty, however, sustaining diversity, and fluidity in low-level co-speech motion likely requires additional mechanisms, such as pose interpolation or higher-level motion representations, particularly for longer continuous sequences.

Across all iterations, participant feedback revealed several recurring themes. Users consistently preferred motion quality over quantity, with expressiveness associated more with smooth, fluid movements than with large or numerous joint activations. Participants were generally more tolerant of under-expressiveness, particularly when unnecessary arm movements were omitted in favor of natural head motions, which may explain why semantically clear categories such as apologies expressed through head bowing were rated more favorably. Synchronization between speech and gesture was also repeatedly emphasized, however, use of Pepper’s native text-to-speech system limited temporal alignment, further encouraging preference for simpler gestures. Finally, participants criticized gestures that appeared semantically unrelated, non-human, or excessively jerky, fast and overpronounced. Collectively, these observations highlight key priorities for improving GenAI-driven co-speech gesture generation using low-level motion primitives.

\section{CONCLUSIONS AND FUTURE WORK}
This work demonstrates that large language models, when paired with iterative reinforcement learning from human feedback (RLHF), meaningfully improve the generation of expressive co‑speech gestures for humanoid robots. By integrating GPT‑4.1 with a function library of low‑level motion primitives and iteratively refining its outputs through five rounds of human evaluation, we show that subjective human preferences can be translated into more expressive, relevant, and fluid robot motion code. Across iterations, user ratings revealed statistically significant improvements in all measured dimensions, with relevance and fluidity benefiting most from the combined effects of few‑shot prompting and direct preference optimization-based finetuning. Qualitative analysis further highlighted how human feedback shaped gesture selection, reduced inappropriate or overly complex motions, and encouraged more semantically aligned behaviors.

At the same time, our findings expose the limitations of relying solely on low‑level function composition for expressive motion generation. While RLHF effectively pruned undesirable behaviors, such as unnecessary and unnatural arm movements, sustaining gesture diversity and achieving smooth, human‑like motion remained challenging, particularly as the model gravitated toward conservative or repetitive patterns (e.g., head nodding). These observations suggest that future systems may benefit from hybrid approaches that combine LLM-driven code generation with higher-level motion abstractions, pose interpolation, movement gradients or learned dynamical models to better capture the richness and nuances of human expressiveness.

Overall, this study provides evidence that human-in-the-loop learning can enhance the dynamic capabilities of LLM-powered robots, offering a scalable path toward more adaptive and human-aligned gesture generation. There is future opportunities of integrating other modalities such as VLM self-critique or visual embodiment, to further ground the LLM. As generative AI continues to advance, integrating human feedback into embodied systems will be essential for developing robots that communicate not only effectively, but expressively and intuitively in real-world interactions.







\section{ACKNOWLEDGMENT}
This research was supported by the Commonwealth through an Australian Government RTP Scholarship [\href{https://doi.org/10.82133/C42F-K220}{DOI: https://doi.org/10.82133/C42F-K220}] and by CSIRO, Australia’s national science organisation, through the Next Generation Graduate Program (NGGP) scholarship. We sincerely thank Takero Izuhara and Tao Lu, former associates from our industry partner ST Solutions PTY LTD, for their ongoing support and ideas, providing research resources, including access to the Pepper robot used in our work.



\bibliographystyle{ieeetr}
\balance

\begin{thebibliography}{10}

\bibitem{zhang_large_2023}
C.~Zhang, J.~Chen, J.~Li, Y.~Peng, and Z.~Mao, ``Large language models for human–robot interaction: {A} review,'' {\em Biomimetic Intelligence and Robotics}, vol.~3, no.~4, p.~100131, 2023.

\bibitem{zhang_generative_2025}
K.~Zhang, P.~Yun, J.~Cen, J.~Cai, D.~Zhu, H.~Yuan, C.~Zhao, and et~al, ``Generative {Artificial} {Intelligence} in {Robotic} {Manipulation}: {A} {Survey},'' 2025.

\bibitem{zeng_large_2023}
F.~Zeng, W.~Gan, Y.~Wang, N.~Liu, and P.~S. Yu, ``Large {Language} {Models} for {Robotics}: {A} {Survey},'' 2023.

\bibitem{wang_large_2024}
J.~Wang, E.~Shi, H.~Hu, C.~Ma, Y.~Liu, X.~Wang, Y.~Yao, and et~al, ``Large language models for robotics: {Opportunities}, challenges, and perspectives,'' {\em Journal of Automation and Intelligence (2025)}, p.~S2949855424000613, 2024.

\bibitem{yoon_speech_2020}
Y.~Yoon, B.~Cha, J.-H. Lee, M.~Jang, J.~Lee, J.~Kim, and G.~Lee, ``Speech gesture generation from the trimodal context of text, audio, and speaker identity,'' {\em ACM Trans. Graph.}, vol.~39, no.~6, pp.~222:1--222:16, 2020.

\bibitem{zabala_modeling_2022}
U.~Zabala, I.~Rodriguez, J.~M. Martínez-Otzeta, and E.~Lazkano, ``Modeling and evaluating beat gestures for social robots,'' {\em Multimed Tools Appl}, vol.~81, no.~3, pp.~3421--3438, 2022.

\bibitem{bartneck_measurement_2009}
C.~Bartneck, D.~Kulić, E.~Croft, and S.~Zoghbi, ``Measurement {Instruments} for the {Anthropomorphism}, {Animacy}, {Likeability}, {Perceived} {Intelligence}, and {Perceived} {Safety} of {Robots},'' {\em Int J of Soc Robotics}, vol.~1, no.~1, pp.~71--81, 2009.

\bibitem{hoffman_primer_2021}
G.~Hoffman and X.~Zhao, ``A primer for conducting experiments in human–robot interaction,'' {\em ACM Transactions of Human-Robot Interaction}, vol.~10, no.~1, 2021.

\bibitem{barmann_incremental_2024}
L.~Bärmann, R.~Kartmann, F.~Peller-Konrad, J.~Niehues, A.~Waibel, and T.~Asfour, ``Incremental learning of humanoid robot behavior from natural interaction and large language models,'' {\em Front. Robot. AI}, vol.~11, 2024.

\bibitem{akalin_reinforcement_2021}
N.~Akalin and A.~Loutfi, ``Reinforcement {Learning} {Approaches} in {Social} {Robotics},'' {\em Sensors}, vol.~21, no.~4, p.~1292, 2021.

\bibitem{kaufmann_survey_2025}
T.~Kaufmann, P.~Weng, V.~Bengs, and E.~Hüllermeier, ``A {Survey} of {Reinforcement} {Learning} from {Human} {Feedback},'' 2025.

\bibitem{ouyang_training_2022}
L.~Ouyang, J.~Wu, X.~Jiang, D.~Almeida, C.~L. Wainwright, P.~Mishkin, {\em et~al.}, ``Training language models to follow instructions with human feedback,'' in {\em Proceedings of the 36th {International} {Conference} on {Neural} {Information} {Processing} {Systems}}, {NeurIPS} '22, (Red Hook, NY, USA), pp.~27730--27744, Curran Associates Inc., 2022.

\bibitem{mahadevan_generative_2024}
K.~Mahadevan, J.~Chien, N.~Brown, Z.~Xu, C.~Parada, F.~Xia, A.~Zeng, and et~al, ``Generative {Expressive} {Robot} {Behaviors} using {Large} {Language} {Models},'' in {\em Proceedings of the 2024 {ACM}/{IEEE} {International} {Conference} on {Human}-{Robot} {Interaction}}, pp.~482--491, 2024.

\bibitem{vemprala_chatgpt_2023}
S.~Vemprala, R.~Bonatti, A.~Bucker, and A.~Kapoor, ``{ChatGPT} for {Robotics}: {Design} {Principles} and {Model} {Abilities},'' 2023.

\bibitem{sobo_evaluating_2025}
A.~Sobo, A.~Mubarak, A.~Baimagambetov, and N.~Polatidis, ``Evaluating {LLMs} for {Code} {Generation} in {HRI}: {A} {Comparative} {Study} of {ChatGPT}, {Gemini}, and {Claude},'' {\em Applied Artificial Intelligence}, vol.~39, no.~1, p.~2439610, 2025.

\bibitem{de_heuvel_impact_2025}
J.~De~Heuvel, D.~Marta, S.~Holk, I.~Leite, and M.~Bennewitz, ``The {Impact} of {VR} and {2D} {Interfaces} on {Human} {Feedback} in {Preference}-{Based} {Robot} {Learning},'' in {\em 2025 {IEEE}/{RSJ} {International} {Conference} on {Intelligent} {Robots} and {Systems} ({IROS})}, pp.~19024--19030, 2025.

\bibitem{pandey_mass-produced_2018}
A.~K. Pandey and R.~Gelin, ``A {Mass}-{Produced} {Sociable} {Humanoid} {Robot}: {Pepper}: {The} {First} {Machine} of {Its} {Kind},'' {\em IEEE Robot. Automat. Mag.}, vol.~25, no.~3, pp.~40--48, 2018.

\bibitem{mehrabian_nonverbal_2017}
A.~Mehrabian, {\em Nonverbal {Communication}}.
\newblock New York: Routledge, 2017.

\bibitem{breazeal_effects_2005}
C.~Breazeal, C.~Kidd, A.~Thomaz, G.~Hoffman, and M.~Berlin, ``Effects of nonverbal communication on efficiency and robustness in human-robot teamwork,'' in {\em 2005 {IEEE}/{RSJ} {International} {Conference} on {Intelligent} {Robots} and {Systems}}, pp.~708--713, 2005.

\bibitem{ende_human-centered_2011}
T.~Ende, S.~Haddadin, S.~Parusel, T.~Wüsthoff, M.~Hassenzahl, and A.~Albu-Schäffer, ``A human-centered approach to robot gesture based communication within collaborative working processes,'' in {\em 2011 {IEEE}/{RSJ} {International} {Conference} on {Intelligent} {Robots} and {Systems}}, pp.~3367--3374, 2011.

\bibitem{lombardi_would_2025}
M.~Lombardi, C.~Calabrese, D.~Ghiglino, C.~Foglino, D.~De~Tommaso, G.~Da~Lisca, L.~Natale, and et~al, ``Would you let a humanoid play storytelling with your child? {A} usability study on {LLM}-powered narrative {Humanoid}-{Robot} {Interaction},'' in {\em 2025 {IEEE}/{RSJ} {International} {Conference} on {Intelligent} {Robots} and {Systems} ({IROS})}, pp.~20066--20073, 2025.

\bibitem{ng-thow-hing_synchronized_2010}
V.~Ng-Thow-Hing, P.~Luo, and S.~Okita, ``Synchronized gesture and speech production for humanoid robots,'' in {\em 2010 {IEEE}/{RSJ} {International} {Conference} on {Intelligent} {Robots} and {Systems}}, pp.~4617--4624, 2010.

\bibitem{moreira_deep_2020}
I.~Moreira, J.~Rivas, F.~Cruz, R.~Dazeley, A.~Ayala, and B.~Fernandes, ``Deep {Reinforcement} {Learning} with {Interactive} {Feedback} in a {Human}–{Robot} {Environment},'' {\em Applied Sciences}, vol.~10, no.~16, p.~5574, 2020.

\bibitem{balsells_autonomous_2023}
M.~Balsells, M.~T. Villasevil, Z.~Wang, S.~Desai, P.~Agrawal, and A.~Gupta, ``Autonomous {Robotic} {Reinforcement} {Learning} with {Asynchronous} {Human} {Feedback},'' in {\em Proceedings of {The} 7th {Conference} on {Robot} {Learning}}, pp.~774--799, PMLR, 2023.

\bibitem{knox_interactively_2009}
W.~B. Knox and P.~Stone, ``Interactively shaping agents via human reinforcement: the {TAMER} framework,'' in {\em Proceedings of the fifth international conference on {Knowledge} capture}, K-{CAP} '09, (New York, NY, USA), pp.~9--16, Association for Computing Machinery, 2009.

\bibitem{kuhlmann_guiding_2004}
G.~Kuhlmann, P.~Stone, R.~Mooney, and J.~Shavlik, ``Guiding a reinforcement learner with natural language advice: {Initial} results in {RoboCup} soccer,'' {\em AAAI Workshop - Technical Report}, 2004.

\bibitem{christiano_deep_2017}
P.~F. Christiano, J.~Leike, T.~B. Brown, M.~Martic, S.~Legg, and D.~Amodei, ``Deep reinforcement learning from human preferences,'' in {\em Proceedings of the 31st {International} {Conference} on {Neural} {Information} {Processing} {Systems}}, {NeurIPS}'17, (Red Hook, NY, USA), pp.~4302--4310, Curran Associates Inc., 2017.

\bibitem{wang_affective_2022}
H.~Wang, J.~Lin, Z.~Ma, Y.~Vasylkiv, H.~Brock, K.~Nakamura, R.~Gomez, and et~al, ``Affective {Behavior} {Learning} for {Social} {Robot} {Haru} with {Implicit} {Evaluative} {Feedback},'' in {\em 2022 {IEEE}/{RSJ} {International} {Conference} on {Intelligent} {Robots} and {Systems} ({IROS})}, pp.~3881--3888, 2022.

\bibitem{juan_shaping_2021}
R.~Juan, J.~Huang, R.~Gomez, K.~Nakamura, Q.~Sha, B.~He, and G.~Li, ``Shaping {Progressive} {Net} of {Reinforcement} {Learning} for {Policy} {Transfer} with {Human} {Evaluative} {Feedback},'' in {\em 2021 {IEEE}/{RSJ} {International} {Conference} on {Intelligent} {Robots} and {Systems} ({IROS})}, pp.~1281--1288, 2021.

\bibitem{jin_data-efficient_2023}
D.~Jin, S.~Mehri, D.~Hazarika, A.~Padmakumar, S.~Lee, Y.~Liu, and M.~Namazifar, ``Data-{Efficient} {Alignment} of {Large} {Language} {Models} with {Human} {Feedback} {Through} {Natural} {Language},'' 2023.

\bibitem{cao_survey_2025}
Y.~Cao, H.~Zhao, Y.~Cheng, T.~Shu, Y.~Chen, G.~Liu, G.~Liang, and et~al, ``Survey on {Large} {Language} {Model}-{Enhanced} {Reinforcement} {Learning}: {Concept}, {Taxonomy}, and {Methods},'' {\em IEEE Transactions on Neural Networks and Learning Systems}, vol.~36, no.~6, pp.~9737--9757, 2025.

\bibitem{wang_reinforcement_2024}
S.~Wang, S.~Zhang, J.~Zhang, R.~Hu, X.~Li, T.~Zhang, J.~Li, and et~al, ``Reinforcement {Learning} {Enhanced} {LLMs}: {A} {Survey},'' {\em arXiv preprint arXiv:2412.10400}, 2024.

\bibitem{yuan_self-rewarding_2024}
W.~Yuan, R.~Y. Pang, K.~Cho, X.~Li, S.~Sukhbaatar, J.~Xu, and J.~Weston, ``Self-rewarding language models,'' in {\em Proceedings of the 41st {International} {Conference} on {Machine} {Learning}}, vol.~235 of {\em {ICML}'24}, (Vienna, Austria), pp.~57905--57923, JMLR.org, 2024.

\bibitem{abeyruwan_i-sim2real_2023}
S.~W. Abeyruwan, L.~Graesser, D.~B. D’Ambrosio, A.~Singh, A.~Shankar, A.~Bewley, D.~Jain, and et~al, ``i-{Sim2Real}: {Reinforcement} {Learning} of {Robotic} {Policies} in {Tight} {Human}-{Robot} {Interaction} {Loops},'' in {\em Proceedings of {The} 6th {Conference} on {Robot} {Learning}}, pp.~212--224, PMLR, 2023.

\bibitem{wang_skill_2022}
X.~Wang, K.~Lee, K.~Hakhamaneshi, P.~Abbeel, and M.~Laskin, ``Skill {Preferences}: {Learning} to {Extract} and {Execute} {Robotic} {Skills} from {Human} {Feedback},'' in {\em Proceedings of the 5th {Conference} on {Robot} {Learning}}, pp.~1259--1268, PMLR, 2022.

\bibitem{stiennon_learning_2020}
N.~Stiennon, L.~Ouyang, J.~Wu, D.~M. Ziegler, R.~Lowe, C.~Voss, A.~Radford, and et~al, ``Learning to summarize from human feedback,'' in {\em Proceedings of the 34th {International} {Conference} on {Neural} {Information} {Processing} {Systems}}, {NeurIPS} '20, (Red Hook, NY, USA), pp.~3008--3021, Curran Associates Inc., 2020.

\bibitem{hiranaka_primitive_2023}
A.~Hiranaka, M.~Hwang, S.~Lee, C.~Wang, L.~Fei-Fei, J.~Wu, and R.~Zhang, ``Primitive {Skill}-{Based} {Robot} {Learning} from {Human} {Evaluative} {Feedback},'' in {\em 2023 {IEEE}/{RSJ} {International} {Conference} on {Intelligent} {Robots} and {Systems} ({IROS})}, pp.~7817--7824, 2023.

\bibitem{tarakli_interactive_2024}
I.~Tarakli, S.~Vinanzi, and A.~D. Nuovo, ``Interactive {Reinforcement} {Learning} from {Natural} {Language} {Feedback},'' in {\em 2024 {IEEE}/{RSJ} {International} {Conference} on {Intelligent} {Robots} and {Systems} ({IROS})}, pp.~11478--11484, 2024.

\bibitem{huang_emotion_2025}
P.~Huang, Y.~Hu, N.~Nechyporenko, D.~Kim, W.~Talbott, and J.~Zhang, ``{EMOTION}: {Expressive} {Motion} {Sequence} {Generation} for {Humanoid} {Robots} {With} {In}-{Context} {Learning},'' {\em IEEE Robotics and Automation Letters}, vol.~10, no.~8, pp.~7699--7706, 2025.

\bibitem{lu_co-speech_2023}
S.~Lu, Y.~Yoon, and A.~Feng, ``Co-{Speech} {Gesture} {Synthesis} using {Discrete} {Gesture} {Token} {Learning},'' in {\em 2023 {IEEE}/{RSJ} {International} {Conference} on {Intelligent} {Robots} and {Systems} ({IROS})}, pp.~9808--9815, 2023.

\bibitem{yazdian_gesture2vec_2022}
P.~J. Yazdian, M.~Chen, and A.~Lim, ``{Gesture2Vec}: {Clustering} {Gestures} using {Representation} {Learning} {Methods} for {Co}-speech {Gesture} {Generation},'' in {\em 2022 {IEEE}/{RSJ} {International} {Conference} on {Intelligent} {Robots} and {Systems} ({IROS})}, pp.~3100--3107, 2022.

\bibitem{roy_gpt-driven_2025}
L.~Roy, E.~A. Croft, A.~Ramirez, and D.~Kulić, ``{GPT}-{Driven} {Gestures}: {Leveraging} {Large} {Language} {Models} to {Generate} {Expressive} {Robot} {Motion} for {Enhanced} {Human}-{Robot} {Interaction},'' {\em IEEE Robotics and Automation Letters}, vol.~10, no.~5, pp.~4172--4179, 2025.

\end{thebibliography}

\end{document}